\documentclass[10pt,twocolumn,letterpaper]{article}

\usepackage[pagenumbers]{cvpr} 

\newcommand{\MODEL}{\textsc{LoCoRe}\xspace}
\newcommand{\custompara}[1]{{\vspace{1mm}\noindent\textbf{#1}\xspace}}

\usepackage{xspace}
\usepackage{multirow} 
\usepackage{hhline} 
\usepackage{tablefootnote}
\usepackage{soul}
\usepackage{arydshln}
\usepackage{enumerate}

\usepackage[utf8]{inputenc} 
\usepackage[T1]{fontenc}    
\usepackage{url}            
\usepackage{booktabs}       
\usepackage{amsfonts}       
\usepackage{nicefrac}       
\usepackage{microtype}      
\usepackage{xcolor}         
\usepackage{siunitx}
\usepackage{mathtools, nccmath}
\usepackage{bbm}
\usepackage{enumerate}

\usepackage{siunitx}
\usepackage{graphicx} 
\usepackage{amsmath}
\usepackage{amssymb}
\usepackage{wrapfig}
\usepackage{float}

\usepackage[accsupp]{axessibility}  

\definecolor{cvprblue}{rgb}{0.21,0.49,0.74}
\usepackage[pagebackref,breaklinks,colorlinks,allcolors=cvprblue]{hyperref}

\def\confName{CVPR}
\def\confYear{2025}

\title{\MODEL: Image Re-ranking with Long-Context Sequence Modeling}

\author{
Zilin Xiao$^{1}$, Pavel Suma$^{2}$, Ayush Sachdeva$^{1}$, Hao-Jen Wang$^{1}$, \\
Giorgos Kordopatis-Zilos$^{2}$, Giorgos Tolias$^{2}$, Vicente Ordonez$^{1}$ \\ \\
$^{1}$Rice University \quad $^{2}$VRG, FEE, Czech Technical University in Prague
}

\begin{document}
\maketitle
\begin{abstract}
We introduce \MODEL, \textbf{Lo}ng-\textbf{Co}ntext \textbf{Re}-ranker, a model that takes as input local descriptors corresponding to an image query and a list of gallery images and outputs similarity scores between the query and each gallery image.
This model is used for image retrieval, where typically 
a first ranking is performed with an efficient similarity measure, and then a shortlist of top-ranked images is re-ranked based on a more fine-grained similarity model.
Compared to existing methods that perform pair-wise similarity estimation with local descriptors or list-wise re-ranking with global descriptors,
\MODEL is the first method to perform list-wise re-ranking with local descriptors.
To achieve this, we leverage efficient long-context sequence models to effectively capture the dependencies between query and gallery images at the local-descriptor level.
During testing, we process long shortlists with a sliding window strategy that is tailored to overcome the context size limitations of sequence models.
Our approach achieves superior performance compared with other re-rankers on established image retrieval benchmarks of landmarks ($\mathcal{R}$Oxf and $\mathcal{R}$Par), products (SOP), fashion items (In-Shop), and bird species (CUB-200) while having comparable latency to the pair-wise local descriptor re-rankers.
\end{abstract}
\section{Introduction}
\label{sec:intro}

\begin{figure*}[t]
	\begin{center}
		\includegraphics[width=\linewidth]{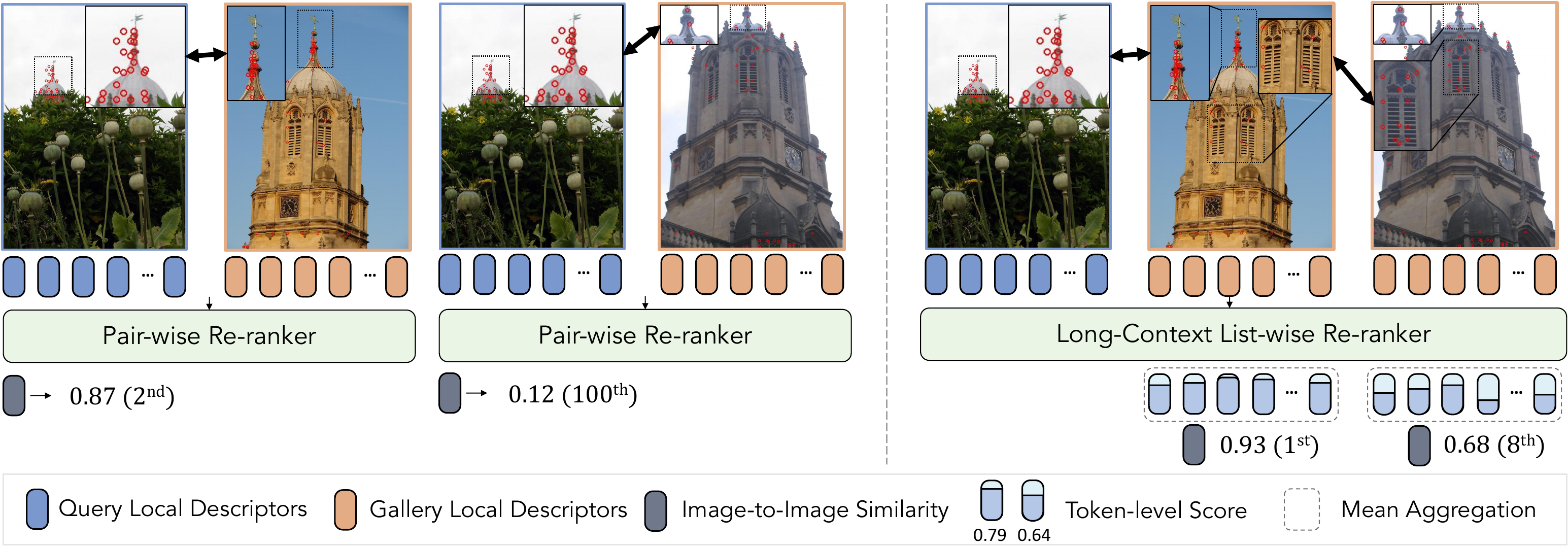}  
	\end{center}
    \vspace{-12pt}
	\caption{\textbf{Overview of pair-wise vs our proposed list-wise re-ranking.}
    Red circles denote the locations of input local descriptors.  
    {\it Left:} The pair-wise re-ranker gets a high score for a positive image since it clearly depicts the same structure at the top of the tower as the query, while a different positive image gets a low score because the top of the tower is not as clearly visible. 
    {\it Right:} Our long-context re-ranker can output a high score for both positive image results since it can exploit the transitive relationship between these images as the two gallery images also share common local descriptors. 
    }
    \vspace{-5pt}
	\label{fig:teaser}
\end{figure*}

Instance-level image retrieval is an important problem in computer vision with many applications. It is usually cast as a problem of metric learning where a model is trained to optimize a distance metric while comparing pairs of examples. Retrieval in large image collections (gallery) is typically performed in two stages. 
First, images are mapped to a compact global image descriptor that is used for efficient retrieval of a shortlist of candidate images from the gallery.
Subsequently, a more powerful but computationally demanding model is used to re-rank the shortlisted images.
Local features and their descriptors are typically employed at this stage to enable a more detailed image-to-image comparison and to provide benefits such as robustness to background clutter and partial visibility due to occlusions.
During re-ranking, the common practice is to estimate the improved similarity in a pair-wise manner by comparing the two local descriptor sets of the query and each image in the shortlist.
Geometric verification is a common re-ranking method where point correspondences are processed with a RANSAC-like process, and a high number of inliers is expected for images depicting the same object under different viewpoints~\cite{objectretrieval2007,DBLP:conf/eccv/CaoAS20}. 
Recent methods train a model while optimizing the image-to-image similarity using either dense~\cite{DBLP:conf/cvpr/LeeS0K22} or sparse~\cite{tan2021instancelevel,DBLP:conf/wacv/ZhangCJZWJ23,suma2024ames,r2former} local representation as its input.
Transformers are becoming a dominant component of such models~\cite{tan2021instancelevel,DBLP:conf/wacv/ZhangCJZWJ23,suma2024ames,r2former}.

Most existing image re-ranking methods perform in a pair-wise way, \ie, the query is separately compared to each of the gallery images.
This strategy does not capture interactions between the gallery images, such as objects or object parts that tend to appear more than once. 
In this work, we introduce \MODEL, a model that performs image re-ranking by jointly processing the input query and an entire shortlist of gallery images at the local descriptor level.
To take into account interactions within the shortlist, we rely on sequence models, in particular transformers, that capture contextual relationships.
Processing within a long context comes with computational challenges.
We overcome those in two ways. 
First, we leverage a model design from existing architectures in Natural Language Processing (NLP) that efficiently extends the maximum context length of a standard transformer.
Second, we propose a sliding window strategy that reuses \MODEL multiple times during inference. 
Figure~\ref{fig:teaser} shows an overview of how the proposed list-wise re-ranker compares to pair-wise re-rankers.

\MODEL takes inspiration from NLP tasks such as sequence tagging~\cite{ramshaw-marcus-1995-text, DBLP:journals/corr/HuangXY15, DBLP:conf/acl/PetersABP17} and extractive question-answering~\cite{DBLP:conf/ecai/EbertsU20, DBLP:conf/emnlp/RajpurkarZLL16, Yoon_2022, DBLP:conf/emnlp/SegalESGB20}. 
Modern solutions to these problems often involve a sequence model that predicts token-level scores to extract task-specific text spans
~\cite{heinzerling-strube-2019-sequence, DBLP:conf/acl/BarbaPN22}. 
In contrast to the current design choice of image re-ranking models,  we adopt the common strategy in NLP and optimize all output tokens, \ie many per gallery image, instead of a single global token. 
At inference time, the aggregated score across tokens associated with a gallery image is used as the similarity value.
By casting image re-ranking as a span extraction task and modeling long context relations across shortlisted images, we demonstrate superior results on established image retrieval benchmarks. 
Specifically, \MODEL is the state-of-the-art re-ranker when evaluated with the same global retrieval method across the CUB-200~\cite{wah_branson_welinder_perona_belongie_2011}, Stanford Online Products (SOP)~\cite{song2016deep} and In-shop~\cite{liuLQWTcvpr16DeepFashion} benchmarks.
Moreover, our models trained on Google Landmarks v2 (GLDv2)~\cite{Weyand_2020_CVPR} are achieving leading results with other re-rankers in relative performance gains on the Revisited Oxford and Paris datasets~\cite{DBLP:conf/cvpr/PhilbinCISZ07, DBLP:conf/cvpr/PhilbinCISZ08, DBLP:conf/cvpr/RadenovicITAC18}.
Code is available at \url{https://github.com/MrZilinXiao/LongContextReranker}.
\section{Related Work}
\label{sec:related_work}

\custompara{Exhaustive search.}
Early image retrieval approaches extract hand-crafted local descriptors that encode visual keypoints in images~\cite{DBLP:journals/ijcv/Lowe04, DBLP:journals/cviu/BayETG08}. The extracted descriptors are compactly quantized into bag-of-words (BoW) representations~\cite{csurka2004bow,sivic2003videogoogle,nister2006vocabtree} to enable exhaustive search in large databases. Aggregating local descriptors into a single global image representation~\cite{DBLP:conf/cvpr/JegouDSP10, sanchez13fisher} allows using simple metrics, such as Euclidean distance, to compute the image-to-image similarity for efficient ranking. Improvements over the vanilla BoW scheme achieve better matching approximation by searching directly with local descriptors~\cite{DBLP:journals/ijcv/ToliasAJ16, jegou2010bow, jegou2008hamming}.

With the rise of deep learning, approaches that leverage neural networks to extract descriptors~\cite{noh2016largescale, DBLP:conf/eccv/CaoAS20, DBLP:conf/cvpr/LeeS0K22, DBLP:journals/corr/abs-2303-10126, arandjelovic2016netvlad} have dominated over the hand-crafted approaches. The main focus is the optimization of the learning process via different loss functions~\cite{patel2022recall, deng2019arcface, schroff2015triplet, hadsell2005contrastive} and network architectures~\cite{DBLP:conf/eccv/CaoAS20, DBLP:conf/cvpr/LeeS0K22, oquab2023dinov2}. In early work~\cite{razavian2016rmatch}, dense image feature maps are extracted from Convolutional Neural Networks (CNN) as local descriptor sets and compared with chamfer similarity. However, aggregating feature maps into a global descriptor remains the dominant approach~\cite{DBLP:conf/iccv/YangHFSXLDH21, babenko2015, tolias2014rmac}. The aggregation can also be learned in an end-to-end fashion via learnable pooling layers~\cite{gempooling2019,psomas2023simpool,superfeatures2022}. Other works~\cite{DBLP:conf/iccv/YangHFSXLDH21, DBLP:conf/eccv/CaoAS20, song2022dalg} demonstrate advantages in training separate global and local descriptor branches, which are ultimately fused into a global descriptor.

\begin{figure*}[h]
    \vspace{5pt}
	\begin{center}
		\includegraphics[width=\linewidth]{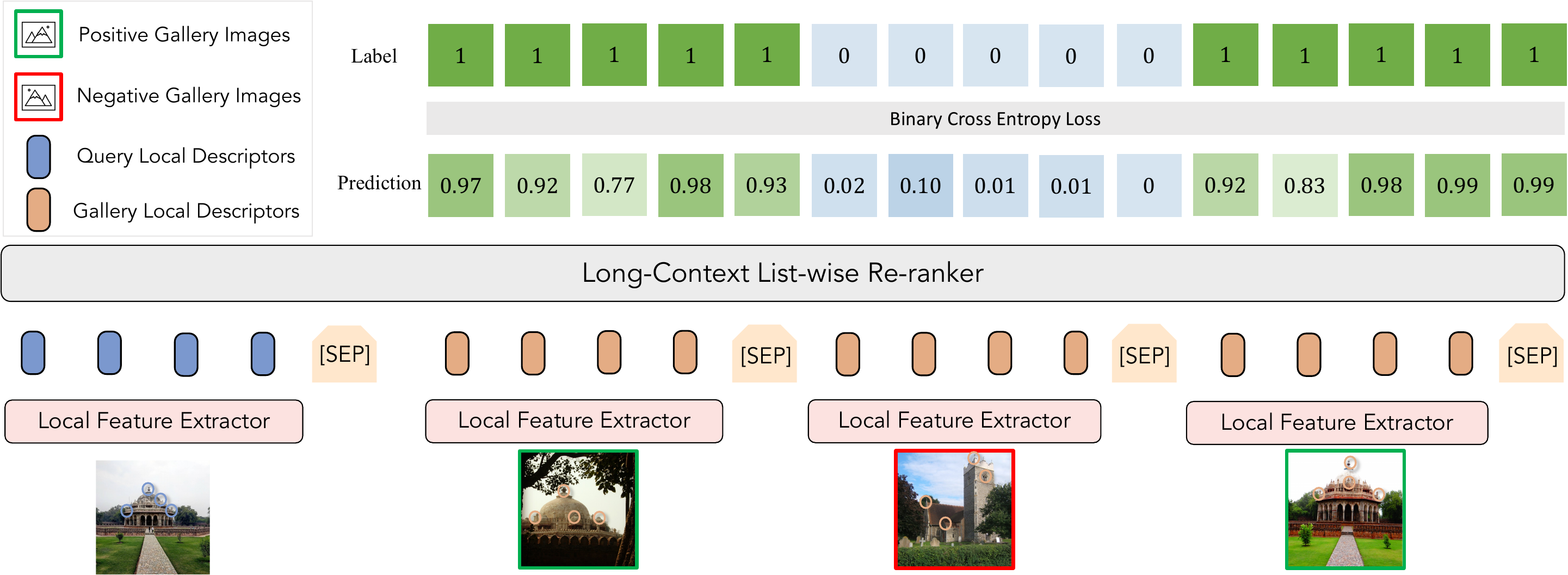}
	\end{center}
    \vspace{-10pt}
	\caption{\textbf{Overview of training and inference under \MODEL when re-ranking three candidate gallery images.} In practice, we re-rank in one inference step up to 100 gallery images. At training time, the model is trained to optimize a binary cross-entropy loss on each gallery image token. At inference time, the token scores of each gallery image get aggregated to facilitate a re-ranked gallery image list. 
    }
    \vspace{-5pt}
	\label{fig:model_train}
\end{figure*}

\custompara{Re-ranking.}
A typical approach for refining the exhaustive search is to apply a re-ranking step that involves more precise similarity estimation. To ensure feasible search times, exhaustive re-ranking is applicable only with global image descriptors. Query expansion derives a new query descriptor by aggregating it with the descriptors of its nearest neighbors from the initial retrieval~\cite{DBLP:conf/iccv/ChumPSIZ07}. Extensions of the base scheme include weighted aggregation based on image ranks~\cite{DBLP:conf/cvpr/ArandjelovicZ12, gempooling2019, Shao_2023_ICCV} or aggregation through a learnable model~\cite{DBLP:conf/eccv/GordoRB20}. An alternative approach is to re-estimate pairwise image similarities by leveraging local descriptors. Due to its high demand in terms of computational complexity, such approaches are only applied to a short list of the top-ranked images provided by the exhaustive search.
Geometric verification (GV) of the descriptor correspondences is a long-standing solution, typically involving RANSAC~\cite{Fischler1981RandomSC}. This approach is applicable to hand-crafted~\cite{DBLP:conf/cvpr/PhilbinCISZ07, DBLP:journals/ijcv/AvrithisT14} as well as learned features~\cite{DBLP:conf/eccv/CaoAS20,simeoni2019local,noh2016largescale}. 
Recently, the geometric approach is surpassed by deep networks optimized to internally match the local descriptors of an image pair and output a single similarity score. Correlation Verification Networks (CVNet)~\cite{DBLP:conf/cvpr/LeeS0K22} incorporates 4D convolutions on top of joint correlations between dense local descriptor sets.
Transformer-based architectures, such as Reranking Transformer (RRT)~\cite{tan2021instancelevel} and Asymmetric and Memory-Efficient
Similarity (AMES)~\cite{suma2024ames}, are designed to be fed with two sets of sparse local descriptors, which are processed as a sequence of tokens to estimate similarity.
A closely related work relying on transformers uses global descriptors to re-rank images in a list-wise manner~\cite{DBLP:conf/nips/OuyangWWZL21} instead of processing each query and gallery image pair separately. It extracts a single affinity image descriptor for each image. Our proposed approach builds on list-wise re-ranking but utilizes local descriptors. This is fundamentally different since each image is represented by multiple descriptors. 
To this end, we repurpose and fine-tune a transformer network specifically designed for long-context tasks.
\section{Method}
In this section, we describe the model architecture, the training procedure and test-time strategies of our proposed \mbox{\MODEL} image reranker.

\subsection{Problem Formulation}   
\label{sec:formulation}
Given a query image $I_q$ and an ordered list of $K$ gallery images $I_{g,i}, i\in\{1,\cdots, K\}$ obtained by global similarity search, 
the purpose of image re-ranking is to produce another refined list of gallery images that are reordered based on improved similarity measures to the query image $I_q$. 

Let $x_q=\left\{\mathbf{x}_{q, j} \in \mathbb{R}^{d}\right\}_{j=1}^L$ denote $L$ local descriptors of dimension $d$ extracted for the query image using a visual backbone and $x_{g,i}=\left\{\mathbf{x}_{g, i, j} \in \mathbb{R}^{d}\right\}_{j=1}^L$ denote the local descriptors for the $i$-th gallery image. 
These $L$ local descriptors are either extracted from a specific layer of a visual backbone model, \eg local descriptors from the last convolutional layer of ResNet~\cite{he2016deep}, or are the top-$L$ multi-scale descriptors weighted by local attention scores, \eg as obtained in DELG~\cite{DBLP:conf/eccv/CaoAS20}.

\custompara{Pair-wise re-ranker.} A typical neural re-ranker $f_\theta$ computes a pair-wise confidence score $S$ for each pair of images $(I_q, I_{g,i})$ based on their local descriptors: 
$$
S(I_q, I_{g, i}) = f_\theta(x_q, x_{g, i}), 
$$
where $f_\theta$ learns a binary classification objective during training to separate matching (positive) and non-matching (negative) image pairs.
Then, the refined gallery list is constructed by sorting the confidence scores of the $K$ gallery images in descending order.
The most common architecture is a transformer~\cite{tan2021instancelevel, suma2024ames}, and the parameter set $\theta$ corresponds to the parameters of the transformer layers and of the binary classifier on top of a single output token.

\custompara{List-wise re-ranker.} 
Instead of estimating the similarity to the query separately per gallery image, we propose a model architecture to perform this jointly across all $K$ images to benefit from interactions across gallery images. 
The deep network takes the local descriptors of the query and $K$ gallery images as input to estimate all $K$ query-to-gallery image similarities via 
$$\mathbf{y} = f_\theta(x_q, x_{g,1}, \ldots, x_{g,K}),$$
where $\mathbf{y} \in [0,1]^K$.
The similarities to perform the sorting and re-ranking are obtained via $S(I_q,I_{g,i})=\mathbf{y}(i)$, with $\mathbf{y}(i)$ being the $i$-th element of vector $\mathbf{y}$.

\subsection{Architecture}

\begin{figure}[t]
    \vspace{3pt}
	\begin{center}
		\includegraphics[width=0.8\linewidth]{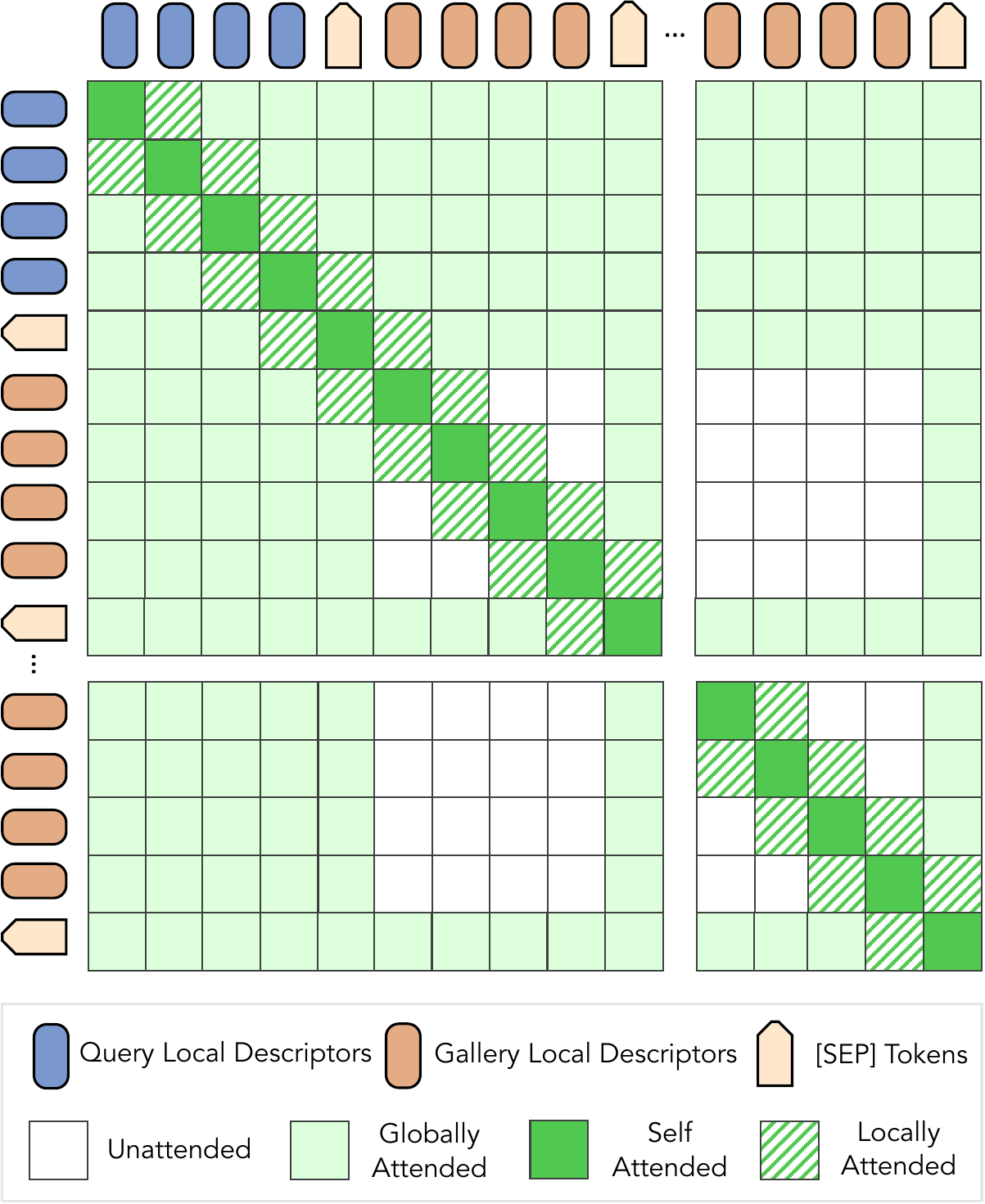}
	\end{center}
    \vspace{-13pt}
	\caption{
 \textbf{Attention pattern of \MODEL} when the number of descriptors for each image is $L=4$ and the local window size is $W=1$. In our experiments we actually use a larger number of local feature descriptors and window size. }
	\label{fig:query_global_attention}
    \vspace{-5pt}
\end{figure}
\custompara{Long context.} We treat the set of input local descriptors as a sequence of vectors by concatenating them.
In principle, any sequence model can serve as a backbone for our method, but to rerank up to a hundred gallery images at once we need a sequence model with a very large context window.
Similar to pair-wise re-rankers, we rely on transformers, but dense attention matrices restrict the model to very small values of $K$ and $L$.
For this reason, we use a pre-trained Longformer~\cite{beltagy2020longformer} as its computational complexity grows linearly with sequence length due to its sliding attention window mechanism. 
It can capture long-range dependencies, as the positive gallery image can appear in any part of the sequence.
Figure~\ref{fig:model_train} presents an overview of our method. 

\custompara{Input sequence.} The vector sequence is given in a matrix format by $x=[x_q, \mathbf{x_s}, x_{g,1},\mathbf{x_s}, \ldots, x_{g,K}, \mathbf{x_s}]$, with $\mathbf{x_s} \in \mathbb{R}^d$ being a learnable separation token, \texttt{[SEP]}, that is meant to act as a global representation of the image that is preceding it. 
In total there are $M=(L+1)(K+1)$ vectors. We use $M$ learnable positional encodings to indicate the position in the sequence of each vector.
Additionally, we use $K+1$ learnable positional encodings to indicate the image from which the vectors come. 
Both these positional encodings are added to the corresponding columns of $x$ and then fed to the sequence model, with each column representing a token.

\custompara{Query global attention. }
Longformer uses a local sliding-window attention layer instead of a full self-attention layer, which allows it to handle longer sequences.
It additionally defines some tokens to perform global attention in a symmetric way, \ie they attend to all other tokens and all other tokens attend to them. 
To compensate for potential issues handling long-range dependencies, we define all tokens associated with the query image and all \texttt{[SEP]} tokens to perform global attention.
In this way, all descriptor tokens attend to other tokens within the same image and of nearby images with a local window size of $W$, while the global attention tokens ensure long-range interactions.
We illustrate the attention pattern in Figure~\ref{fig:query_global_attention}. 
Note that using global attention for the limited set of query and separation tokens brings only a marginal computation overhead.
Additionally, such an attention mechanism is an important inductive bias, without which the model demonstrates trivial re-ranking performance, as indicated in our later experiments.

\custompara{Token classifier. } 
All contextualized output representations are fed to a binary classifier, which aims to predict whether the corresponding input token comes from a positive or a negative image.
This resembles the design in a variety of natural language processing tasks that require token-level classification, such as named entity recognition~\cite{DBLP:conf/naacl/BarbaPN21, DBLP:conf/acl/BarbaPN22}, parts-of-speech tagging~\cite{nguyen2016robust}, and extractive question answering~\cite{DBLP:conf/emnlp/RajpurkarZLL16, Yoon_2022, DBLP:conf/emnlp/SegalESGB20}.

\subsection{Training}  
\label{sec:model_training}
During training, all $(L+1) \times K$ classifier outputs are fed to a binary cross-entropy loss according to the ground-truth labels of query-gallery-image pairs.
Regarding training batch sampling, we pick a query image from the training set, use global similarity search, and pick the top-$K$ training images as gallery images to form a list-wise training sample.
However, directly training \MODEL on list-wise re-ranking supervision comes with inherent issues. 
Global similarity search tends to have positive gallery images ranked at the top positions of the retrieved gallery list. 
Feeding the local descriptors for the gallery images in the order given by the global retrieval model introduces a rather counterproductive positional bias. 
Models can use this bias as a shortcut during training since the model will correlate the positive matching predictions with top positions rather than relying on learning to match local descriptors. 
Therefore, we shuffle the gallery list in each training step to ensure that each position has an equal probability of being assigned a positive gallery image. We refer to this strategy as {\em gallery shuffled training}. 

\subsection{Inference}  

\custompara{Image-to-image similarity.} 
During test time, we compute a single score per gallery image by aggregating all $L+1$ classifier outputs that are associated with the specific image. The aggregator function can be defined as one of the following: (i) \textit{the average token score}, (ii) \textit{the first token score} or (iii) \textit{\texttt{[SEP]} token score} of the token span associated with the corresponding gallery images.

\begin{figure}[t]
	\begin{center}
        \includegraphics[width=\linewidth]{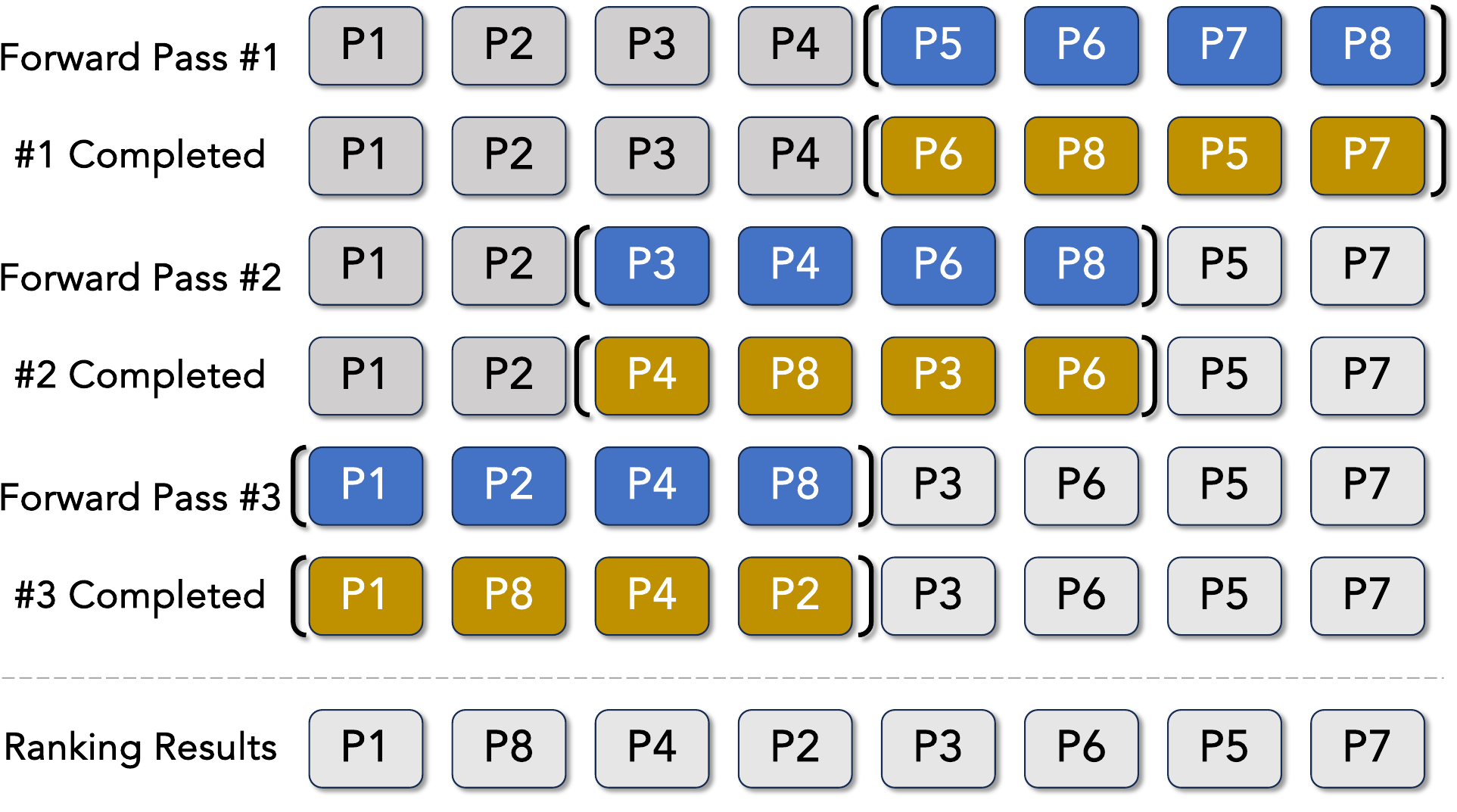}
	\end{center}
    \vspace{-14pt}
	\caption{\textbf{Illustration of sliding window re-ranking} for $N=8$ gallery images with a list-wise re-ranker that can re-rank $K=4$ images each forward pass. Blue blocks represent the re-ranking window of the current forward pass, and it slides to the next window with a stride of $S=2$ images. Brown blocks indicate the re-ranking for the current forward pass is completed. 
    }
	\label{fig:sliding_window}
    \vspace{-5pt}
\end{figure}

\custompara{Sliding window re-ranking.} 
Despite using a long-context sequence model, we might still be limited in how many gallery images we can rerank at once. In order to extend the number of gallery images at inference time, we can rerank separate overlapping groups of gallery images to obtain reranking scores for a larger number of gallery images.
Figure~\ref{fig:sliding_window} illustrates our proposed test-time sliding window re-ranking strategy.
Let $N$ be the total number of gallery images pending for re-ranking and $K$ be the maximum number of images that the reranker can process at once. We first use the model to rank the $(N-K)$-th to $N$-th gallery images and slide the re-ranking window of size $K$ forward with a stride step size of $S$. This process continues to the start of the shortlist to maximally ensure each gallery image gets to its ranking as accurately as possible. Our experiments demonstrate consistent improvements under this strategy.

\section{Experiments}
We first introduce the experimental setup in Section~\ref{sec:exp_setup} which covers datasets, metrics, and details of training and evaluation. 
We report the main results and ablation results in Section~\ref{sec:main_results} and Section~\ref{sec:ablation}.

\subsection{Experimental Setup}
\label{sec:exp_setup}

\custompara{Datasets and Metrics. }
To demonstrate the effectiveness of our method, we experiment with large-scale instance-level recognition datasets. We train on the clean Google Landmarks v2 (GLDv2) dataset~\cite{Weyand_2020_CVPR} and evaluate on the Revisited Oxford ($\mathcal{R}$Oxf) and Paris ($\mathcal{R}$Par) datasets and their 1M distractor variants ($\mathcal{R}$Oxf+1M, $\mathcal{R}$Par+1M)~\cite{DBLP:conf/cvpr/PhilbinCISZ07, DBLP:conf/cvpr/PhilbinCISZ08, DBLP:conf/cvpr/RadenovicITAC18}. Following the literature in the relevant field~\cite{tan2021instancelevel,suma2024ames}, we report mean average precision (mAP) on the Medium and Hard settings.
In addition, following established image retrieval benchmarks, we evaluate on the following datasets: CUB-200~\cite{wah_branson_welinder_perona_belongie_2011}, Stanford Online Products (SOP)~\cite{song2016deep} and In-shop~\cite{liuLQWTcvpr16DeepFashion}. 
We choose Recall@$k$ (R@$k$) and mean average precision at $R$ (mAP@$R$) as evaluation metrics for these datasets, which are the standard metrics~\cite{song2016deep,liuLQWTcvpr16DeepFashion,tan2021instancelevel}. 

\begin{table*}[t]
\small
\renewcommand{\arraystretch}{1.2}
\centering
{
\begin{tabular}{l @{\hspace{5pt}}c c c c c c c c c c}
\toprule
 \small \multirow{3}{*}{\bf Method} & \multirow{3}{*}{\shortstack{\#  \\ {\bf desc.}}} & \multicolumn{4}{c}{ \bf\small Medium} && \multicolumn{4}{c}{\bf\small Hard\footnotemark{}} \\[+0.3em] \cmidrule[0.5pt]{3-6} \cmidrule[0.5pt]{8-11}
 & & \multicolumn{1}{c}{\large \vphantom{M} \scriptsize $\mathcal{R}$Oxf } & \multicolumn{1}{c}{\scriptsize $\mathcal{R}$Oxf+1M} & \multicolumn{1}{c}{\scriptsize $\mathcal{R}$Par} & \multicolumn{1}{c}{\scriptsize $\mathcal{R}$Par+1M} && \multicolumn{1}{c}{\scriptsize $\mathcal{R}$Oxf} & \multicolumn{1}{c}{\scriptsize $\mathcal{R}$Oxf+1M} & \multicolumn{1}{c}{\scriptsize $\mathcal{R}$Par} & \multicolumn{1}{c}{\scriptsize $\mathcal{R}$Par+1M}  \\

\midrule
\multicolumn{11}{c}{\small DELG global \& local}\\\midrule
RN50-DELG \cite{DBLP:conf/eccv/CaoAS20} & - & \num{73.6} & \num{60.6} & \num{85.7} & \num{68.6} &&  \num{51.0} & \num{32.7} & \num{71.5} & \num{44.4}\\
\quad + GV~\cite{DBLP:conf/eccv/CaoAS20} & 1000 & \num{78.3} & \num{67.2} & \num{85.7} & \num{69.6} &&  \num{57.9} & \num{43.6} & \num{71.0} & \num{45.7} \\
\quad + RRT~\cite{tan2021instancelevel} & 500 & \num{78.1} & \num{67.0} & \num{86.7} & \num{69.8} &&  \num{60.2} & \num{44.1} & \num{75.1} & \num{49.4} \\
\quad + CVNet Reranker~\cite{DBLP:conf/cvpr/LeeS0K22} & 3,072 & \num{78.7} & \num{67.7} & \num{87.9} & \num{72.3} &&  \num{63.0} & \num{46.1} & \num{76.8} & \num{52.5} \\
\quad + \MODEL-small & 50 & \num{80.9} & \num{70.9} & \textbf{\num{89.3}} & \num{77.7} && \num{63.3} & \num{49.3} & \textbf{\num{77.8}} & \num{57.9} \\
\quad + \MODEL-base & 50 & \textbf{\num{81.6}} & \textbf{\num{71.7}} & \num{89.2} & \textbf{\num{77.9}} && \textbf{\num{64.2}} & \textbf{\num{50.5}} & \num{77.6} & \textbf{\num{58.2}} \\
\midrule
RN101-DELG~\cite{DBLP:conf/eccv/CaoAS20} & - & \num{76.3} & \num{63.7} & \num{86.6} & \num{70.6} &&  \num{55.6} & \num{37.5} & \num{72.4} & \num{46.9} \\

\quad + GV~\cite{DBLP:conf/eccv/CaoAS20} & 1000 &  \num{81.2} & \num{69.1} & \num{87.2} & \num{71.5} &&  \num{64.0} & \num{47.5} & \num{72.8} & \num{48.7} \\

\quad + RRT~\cite{tan2021instancelevel} & 500 & \num{79.9} & - & \num{87.6} & - &&  \num{64.1} & - & \num{76.1} & - \\
\quad + \MODEL-small & 50 & \num{81.8} & \num{74.1} & \num{87.9} & \num{75.9} && \num{64.2} & \num{54.7} & \num{75.0} & \num{55.2} \\

\quad + \MODEL-base & 50 & \textbf{\num{83.3}} & \textbf{\num{76.1}} & \textbf{\num{90.3}} & \textbf{\num{80.7}} && \textbf{\num{66.9}} & \textbf{\num{56.2}} & \textbf{\num{81.4}} & \textbf{\num{61.8}} \\

\midrule
\multicolumn{11}{c}{\small SuperGlobal global \& DINOv2 local}\\\midrule


RN101-SuperGlobal~\cite{Shao_2023_ICCV} & - & \num{85.3} & \num{78.8} & \num{92.1} & \num{83.9} && \num{72.1} & \num{61.9} & \num{83.5} & \num{69.1} \\  

\quad + AMES~\cite{suma2024ames} & 50 & \num{86.4} & \num{80.3} & \num{91.7} & \num{83.7} && \num{73.8} & \num{65.4} & \num{83.6} & \num{70.9} \\
\quad + AMES$^\ast$~\cite{suma2024ames} & 50 & \num{86.4} & \num{80.4} & \num{91.7} & \num{83.8} && \num{74.4} & \num{65.7} & \num{83.7} & \num{71.3} \\
\quad + \MODEL-small & 50 & \num{89.0} & \num{82.7} & \num{91.9} & \num{84.0} && \num{77.6} & \num{68.9} & \num{84.9} & \num{72.3} \\
\quad + \MODEL-base & 50 & \num{89.7} & \num{83.6} & \num{92.1} & \num{84.3} && \num{79.5} & \num{71.2} & \num{85.2} & \num{72.7} \\
\cmidrule{2-11}
\quad + AMES (top-400)~\cite{suma2024ames} & 50 & \num{87.4} & \num{81.2} & \num{92.8} & \num{85.7} && \num{75.2} & \num{67.0} & \num{85.6} & \num{73.9} \\
\quad + AMES$^\ast$ (top-400)~\cite{suma2024ames} & 50 & \num{87.4} & \num{81.3} & \num{92.8} & \num{85.8} && \num{75.6} & \num{67.3} & \num{85.8} & \num{74.3} \\
\quad + SuperGlobal Rerank (top-400)~\cite{Shao_2023_ICCV} & - & \num{90.9} & \num{84.4} & \num{93.3} & \num{84.9} && \num{80.2} & \num{71.1} & \num{86.7} & \num{71.4} \\   
\quad + \MODEL-small (top-400$^\dagger$) & 50 & \num{91.0} & \num{84.7} & \num{93.3} & \num{86.4} && \num{79.9} & \num{70.9} & \num{87.3} & \num{75.8} \\

\quad + \MODEL-base (top-400$^\dagger$) & 50 & \textbf{\num{92.0}} & \textbf{\num{85.8}} & \textbf{\num{93.8}} & \textbf{\num{86.8}} && \textbf{\num{81.8}} & \textbf{\num{73.2}} & \textbf{\num{87.7}} & \textbf{\num{76.5}} \\

\bottomrule
\end{tabular}
}
\vspace{-5pt}
\caption{
\textbf{Performance (mAP) on $\mathcal{R}$Oxf and $\mathcal{R}$Par} and their 1M distractor variants (+1M) with Medium and Hard evaluation strategy. For a fair comparison, results for re-rankers are reported with their top-$100$ candidates unless indicated otherwise. 
Re-ranking for RN50-DELG and RN101-DELG are with DELG local descriptors while re-ranking for RN101-SuperGlobal is based on DINOv2-ViT-B/14~\cite{oquab2023dinov2} local descriptors. $^\ast$ indicates AMES trained with the hidden size of 768, serving as a fair comparison with \MODEL. 
\linebreak$^\dagger$~indicates sliding window re-ranking is enabled with a stride size $S=50$. 
}
\label{tab:sota_main}
\vspace{-10pt}
\end{table*}

\custompara{Training and Evaluation. } 
Unless stated otherwise, all experiments follow the configuration of $L=50$ and $K=100$, \ie re-ranking is conducted with 1 query image and 100 gallery images, each of them providing 50 descriptors. In inference, we use the \texttt{[SEP]} token score as the default choice to get individual image similarity scores.
For each dataset, we follow the official training and validation splits if available. Otherwise, we consider the first half of the training dataset as the training split and the remaining as the validation. Hyper-parameters are tuned based on the validation split. 
We utilize a series of models ranging from 19.4M (tiny) to 111.8M (base) parameters. 
The implementation details employed for each dataset are reported in the supplementary material.

\custompara{Competing Methods.}
To verify the versatility of our method on the landmarks datasets, we compare \MODEL with state-of-the-art re-ranker approaches. We evaluate with local descriptors extracted from two different backbones, \ie DELG~\cite{DBLP:conf/eccv/CaoAS20} and DINOv2~\cite{oquab2023dinov2}.
For the DELG experiments, 
we follow the original process to extract global and local descriptors at 3 scales ($\left\{1 / \sqrt{2}, 1, \sqrt{2}\right\}$). The top-$L$ local descriptors are selected for each image based on the provided DELG weights. We compare with the RRT~\cite{tan2021instancelevel} and CVNet Reranker~\cite{DBLP:conf/cvpr/LeeS0K22} which report results on top of the same descriptors. 
We further present the results of GV~\cite{objectretrieval2007,DBLP:conf/eccv/CaoAS20} as reported in the original DELG paper. 

In addition, we experiment with local descriptors from DINOv2-ViT-B/14~\cite{oquab2023dinov2}. 
We follow the descriptor extraction of AMES~\cite{suma2024ames} to extract descriptors and to be directly comparable to this re-ranking method.
We also adopt their global-local ensemble scheme and combine the local similarities from \MODEL with the global similarities. In these experiments, we extract global descriptors using the SuperGlobal~\cite{Shao_2023_ICCV} ResNet-101 backbone. Our ensemble parameters are tuned on the public test split of the GLDv2 dataset in the same manner as AMES. We compare \MODEL with the publicly-available AMES model. For a fair comparison, we also train AMES with descriptors of $d=768$ dimensionality,  \ie same as \MODEL, using the
official implementation. 

Regarding the experiments on the other image retrieval benchmarks,
we follow previous work~\cite{tan2021instancelevel} and extract global and local descriptors based on a CNN backbone~\cite{DBLP:conf/nips/OuyangWWZL21, DBLP:conf/nips/ShenXHSA21}.
More precisely, we use ProxyNCA++~\cite{teh2020proxynca} to train a ResNet-50 backbone with randomly cropped $224 \times 224$ images, which is used to extract local descriptors from the dense feature maps of the last convolutional layer. Then, the channel dimension of the feature maps is reduced to the input dimension of re-rankers using a $1 \times 1$ convolutional layer.
On SOP, we directly compare with the results reported in RRT~\cite{tan2021instancelevel}. For CUB-200 and InShop, we train RRT with the provided implementation and in the identical setup. Furthermore, we compare with re-rankers that operate with global descriptors, \ie SSR Rerank~\cite{DBLP:conf/nips/ShenXHSA21}, AQE~\cite{DBLP:conf/iccv/ChumPSIZ07}, DQE~\cite{DBLP:conf/cvpr/ArandjelovicZ12}, and $\alpha$QE~\cite{DBLP:journals/pami/RadenovicTC19}. 
Evaluation for these datasets follows the same evaluation protocols as {\em Jun~et.al.}~\cite{DBLP:journals/corr/abs-1903-10663}.

\footnotetext{indicates the hard setting allows {\em easy} images to be used in the re-ranking and removed before evaluation following~\cite{Shao_2023_ICCV}. Details in the supplementary clarify how such differences impact the final results.}

\begin{table*}[t]
    \centering
    \setlength{\tabcolsep}{2.4pt}
    \renewcommand{\arraystretch}{1.1}
    \small
    \begin{tabular}{lcccc c cccc c cccc}
    
    \toprule
    \textbf{} & \multicolumn{4}{c}{\textbf{CUB-200}} && \multicolumn{4}{c}{\textbf{SOP}} && \multicolumn{4}{c}{\textbf{In-Shop}} \\
    \cmidrule{2-5}\cmidrule{7-10}\cmidrule{12-15}
    & R$@1$ & R$@2$ & R$@4$ & mAP && R$@1$ & R$@10$ & R$@100$ & mAP && R$@1$ & R$@10$ & R$@20$ & mAP \\
    \midrule
    Global descriptors & 
    68.9 & 79.4 & 87.3 & 49.8 && 
    80.8 & 92.1 & 96.9 & 65.1 &&
    88.5 & 97.7 & 98.4 & 74.8 \\
    \midrule
    
    SSR Rerank \cite{DBLP:conf/nips/ShenXHSA21} & 
    69.4 & 79.0 & 86.1 & 54.2 &&
    81.2 & 91.9 & 95.6 & 66.7 &&
    86.3 & 97.1 & 98.2 & 75.8 \\
    
    AQE \cite{DBLP:conf/iccv/ChumPSIZ07} &
    66.9 & 76.8 & 82.7 & 58.4 && 
    76.9 & 89.3 & 94.5 & 66.1 &&
    81.7 & 95.9 & 96.7 & 73.2 \\ 
    
    DQE \cite{DBLP:conf/cvpr/ArandjelovicZ12} &
    67.0 & 75.5 & 82.0 & 54.6 &&
    67.9 & 81.5 & 90.6 & 47.8 &&
    86.4 & 97.1 & 98.1 & 72.4 \\
    
    $\alpha$QE \cite{DBLP:journals/pami/RadenovicTC19} &
    70.9 & 78.8 & 84.7 & 56.9 &&
    81.1 & 90.7 & 96.3 & 68.1 &&
    88.5 & 97.1 & 98.1 & 76.8 \\
    \midrule
    
    RRT \cite{tan2021instancelevel} & 
    68.7 & 85.0 & 95.6 & 55.6 &&
    81.9 & 92.4 & \textbf{96.9} & 67.2 &&
    88.3 & \textbf{97.9} & 98.6 & 77.6 \\
    
    \MODEL-tiny &
    71.4 & 86.8 & 96.4 & 58.1 &&
    82.4 & \textbf{93.1} & \textbf{96.9} & 68.0 &&
    89.1 & \textbf{97.9} & 98.2 & 78.4 \\
    
    \MODEL-small &
    74.6 & 89.1 & 97.3 & 61.0 &&
    83.3 & 92.7 & \textbf{96.9} & 69.4 &&
    \textbf{89.4} & 97.7 & 97.7 & \textbf{78.8} \\
    
    \MODEL-base &
    \textbf{78.3} & \textbf{91.9} & \textbf{98.2} & \textbf{64.8} &&
    \textbf{83.8} & 92.9 & \textbf{96.9} & \textbf{71.0} &&
    87.9 & \textbf{97.9} & \textbf{98.7} & 77.0 \\
    
    \bottomrule
    \end{tabular}
    \vspace{-5pt}
    \caption{\textbf{Performance (R$@k$ and mAP) on metric learning benchmarks}. Results are reported on the same sets of global descriptors to showcase the effectiveness of re-rankers fairly. mAP refers to the mAP@$R$. Re-ranking 100 images is used.
    }
    \label{tab:small_benchmark_reranking_2}
    \vspace{-5pt}
\end{table*}

\subsection{Results}
\label{sec:main_results}
\custompara{Retrieval on landmarks.}
We report \MODEL performance on $\mathcal{R}$Oxf and $\mathcal{R}$Par in Table~\ref{tab:sota_main}. 
We observe that \MODEL-base exhibits clear and consistent improvement over all local-descriptor re-ranking baselines, using different types of descriptors as input and across all settings. The improvement is most pronounced on $\mathcal{R}$Oxf+1M and $\mathcal{R}$Par+1M in the hard setting, where it achieves significant gains of +17.8 and +13.8 mAP points on RN50-DELG, against the prior state-of-the-art CVNet-Reranker, which achieved +13.4 and +8.1 mAP points, respectively. 
The base model shows a significant boost compared to the small one.
We outperform even the latest state-of-the-art of pair-wise models while re-ranking either 100 or 400 images. The proposed sliding window re-ranking allows us to effectively go beyond the list size used during training.

\custompara{Retrieval on metric learning benchmarks.}
We report \MODEL performance on CUB-200~\cite{wah_branson_welinder_perona_belongie_2011}, SOP~\cite{song2016deep} and In-Shop~\cite{liuLQWTcvpr16DeepFashion} and compare with previous re-ranking methods in Table~\ref{tab:small_benchmark_reranking_2}. 
Query expansion methods and SSR Rerank lead to marginal or no improvements in retrieval performance, yet we observe inconsistent behaviour across datasets, \eg, the DQE method on the SOP dataset severely compromises the original performance of the global retriever. 
RRT demonstrates improvements across datasets. 
The re-ranking performance of \MODEL variants is consistently robust, as it shows significant improvements across all datasets and is aligned with increases in both R@$k$ and mAP@$R$. 
This is particularly evident from the relative improvements observed on the CUB-200 dataset, where \MODEL-base obtains a +9.4 increase in R@1 and a +15.0 increase in mAP@$R$, as opposed to the slight decrease of -0.2 in R@1 and smaller gain of +5.8 in mAP@$R$ for RRT.
Notably, this is the first work that reports performance improvements with local descriptors re-ranking on CUB and In-Shop. The only previous work reporting such results on SOP is RRT.

\begin{table}[t]
\small
\centering
\begin{tabular}{c c @{\hspace{20pt}} c c c @{\hspace{-3pt}} c c}
\toprule
\multirow{2}{*}{$N$} & \multirow{2}{*}{\bf $S$} & \multicolumn{2}{c}{\bf Medium} && \multicolumn{2}{c}{\bf Hard} \\ \cmidrule[0.5pt]{3-4} \cmidrule[0.5pt]{6-7}
& & \scriptsize $\mathcal{R}$Oxf+1M & \scriptsize $\mathcal{R}$Par+1M && \scriptsize $\mathcal{R}$Oxf+1M & \scriptsize $\mathcal{R}$Par+1M \\
\midrule
 -  &  -  & \num{78.5} & \num{83.6} && \num{61.4} & \num{68.4} \\
\midrule
100 & N/A & \num{82.7} & \num{84.0} && \num{68.9} & \num{72.3} \\
200 & 50  & \num{83.9} & \num{85.1} && \num{70.4} & \num{74.3} \\
400 & 25  & \textbf{\num{84.8}} & \textbf{\num{86.4}} && \num{70.8} & \textbf{\num{75.9}} \\
400 & 50  & \num{84.7} & \textbf{\num{86.4}} && \textbf{\num{70.9}} & \num{75.8} \\
400 & 75  & \num{84.5} & \num{86.1} && \num{70.7} & \num{75.7} \\
400 & 100 & \num{82.9} & \num{84.9} && \num{69.1} & \num{73.1} \\
\bottomrule
\end{tabular}
\vspace{-5pt}
\caption{\textbf{Different settings of our sliding window strategy} for re-ranking with \MODEL-small and DINOv2 local descriptors. One forward pass has length of $K=100$ for each setting. The sliding windows are used with a stride of $S$ images, and $N$ images are re-ranked in total. \label{tab:sliding}}
\vspace{-5pt}
\end{table}

\subsection{Ablation Study and Analysis}
\label{sec:ablation}

\begin{table}[t]
\centering
\small
\setlength{\tabcolsep}{3pt}
\renewcommand{\arraystretch}{1.1}
\begin{tabular}{l @{\hspace{10pt}} lll}
\toprule
\textbf{Ablation Module} & R$@1$ & R$@10$ & mAP \\
\midrule
Global descriptors & 80.8 & 92.1 & 65.1 \\
\midrule
RRT~\cite{tan2021instancelevel} & 81.9 & 92.4 & 67.2 \\  
\midrule
\MODEL-tiny & 82.4 & 93.1 & 68.0 \\
\hspace{0.5em} w/o \textit{gallery shuffled training} & 80.7 & 91.9 & 65.1 \\
\hspace{0.5em} w/o \textit{global query attention} & 60.7 & 90.0 & 53.0 \\
\hspace{0.5em} w/ ~\textit{shuffled evaluation} & 81.8 \tiny{$\pm\text{0.1}$} & 92.8 \tiny{$\pm\text{0.1}$} & 67.5 \tiny{$\pm\text{0.0}$} \\
\midrule
\MODEL-small & 83.3 & 92.7 & 69.4 \\
\midrule
\MODEL-base & 83.8 & 92.9 & 71.0 \\
\hspace{0.5em} w/ \textit{first token score} & 83.7 & 92.5 & 70.9 \\
\hspace{0.5em} w/ \textit{\texttt{[SEP]} token score} & 83.7 & 92.5 & 70.9  \\
\bottomrule
\end{tabular}
\vspace{-5pt}
\caption{\textbf{Ablation studies for \MODEL} on the SOP dataset. mAP refers to the mAP@$R$. Re-ranking 100 images is used.
}
\label{tab:ablation_sop}
\vspace{-5pt}
\end{table}

\begin{table*}[t]
\small
\centering
\setlength\tabcolsep{5pt}
\renewcommand{\arraystretch}{1.2}
\begin{tabular}{llllllllll}
    \toprule
   \multirow{2}{*}{\textbf{Models}} && \multicolumn{3}{c}{\MODEL} & \multirow{2}{*}{RRT~\cite{tan2021instancelevel}} & \multirow{2}{*}{CVNet~\cite{DBLP:conf/cvpr/LeeS0K22}} & \multirow{2}{*}{AMES$^\ast$~\cite{suma2024ames}}  \\ \cmidrule{2-5}
   && tiny & small & base & & & &  \\
   \midrule
   \textbf{\# of Params} && 19.4M & 58.7M & 111.8M & 2.2M & 7.5M & 32.1M\\
   \textbf{FLOPs} && 200.0G & 517.5G & 1035.1G& 518.5G & 2087.0G & 679.2G \\
   \textbf{FLOP/s (T)} && 10.8${_{\pm\text{0.1}}}$ & 20.9${_{\pm\text{0.2}}}$ & 7.4${_{\pm\text{0.2}}}$ & 7.0${_{\pm\text{0.2}}}$ & 1.8${_{\pm\text{0.3}}}$ & 11.1${_{\pm\text{0.0}}}$ \\
   \textbf{Latency (ms)} && 18.5${_{\pm\text{0.2}}}$ & 24.7${_{\pm\text{0.3}}}$ & 140.1${_{\pm\text{4.2}}}$ & 74.4${_{\pm\text{4.5}}}$ & 1418${_{\pm\text{323}}}$ & 61.1${_{\pm\text{0.5}}}$ \\
   \textbf{Peak Memory} && 3.2GB & 4.5GB & 8.9GB & 29.0GB & 130.0MB & 6.0GB \\
   \bottomrule
\end{tabular}
\vspace{-5pt}
\caption{\textbf{Runtime performance comparison of \MODEL variants with other re-ranking methods}. Floating-point operation (FLOP), latency and memory consumption are measured per 100 re-ranked gallery images with a single query image. $^\ast$ indicates AMES trained with the hidden size of $768$. RRT and CVNet are measured in their original settings.}
\label{tab:runtime}
\vspace{-5pt}
\end{table*}

\custompara{Sliding window re-ranking with \MODEL.}
Table~\ref{tab:sliding} presents retrieval performance for different settings of the sliding window approach. This approach effectively re-uses \MODEL to extend the list size defined during training, which is according to the hardware limitations, \ie the GPU memory. 
We observe that re-ranking with $N>K$ increases performance compared to $N=K$ and also that a smaller stride is better up to some extent. 
We also find that provided $S<K$, \ie overlapping the re-ranking window in each sliding process, greatly improves performance, with the performance gains saturating as $S$ further decreases. This is consistent with our motivation in designing the sliding window re-ranking -- to promote interaction across the gallery within the shortlist beyond the context window of the sequence model.
As a compromise between performance and runtime, $S=50$ is used for the experiment in Table~\ref{tab:sota_main}.

\custompara{Training and evaluation strategies.} 
Table~\ref{tab:ablation_sop} shows the results of ablation studies on the SOP benchmark. Without \textit{gallery shuffled training}, \MODEL-tiny exhibits trivial results that are nearly identical to global retrieval. 
This is because the model learns from positional shortcut as discussed in Section~\ref{sec:model_training} and simply repeats the global retrieval results instead of modeling list-wise correspondences. We also observe a significant performance decline when disabling \textit{global query attention}, suggesting that the absence of global query attention makes the model suffer from learning collapse due to challenges in capturing long-distance correspondences.

To investigate whether \MODEL takes advantage of candidate confidence permutations in global retrieval, \ie, the re-ranker preferentially assigns higher scores to top retrieved candidates, we conduct an ablation with \textit{shuffled evaluation}, where we shuffle the candidates in the gallery list even at inference time. We repeat the experiment using 5 different random seeds and report the mean and the standard deviation. Reduced scores demonstrate that the re-ranker has learned to use candidate confidence permutation from global retrieval results, a signal that pair-wise re-rankers cannot exploit.

Finally, we investigate the impact of different strategies to aggregate the individual scores of all tokens of each image. The results with the base model indicate that various aggregators have no substantial impact on re-ranking performance, reflecting the robustness of the proposed method.

\custompara{Scalability. }
Table~\ref{tab:runtime} shows that the previous largest neural re-ranker CVNet only has 7.5M parameters, in contrast with our smallest model, \MODEL-tiny, which has 19.4M. 
To explore the potential for scaling up our model, we conduct ablations on model size. 
The ablation results demonstrate that model performance improves consistently with increases in model size, indicating considerable potential for scalability in our long-context re-ranking method. 
On the other hand, scaling up pair-wise re-rankers, like RRT, does not yield notable performance improvements. Furthermore, we discuss in the following section a substantial computational cost introduced by large pair-wise re-rankers.

\custompara{Runtime Performance.}
One major advantage of our approach is that \MODEL can complete the top-$k$ re-ranking with a single forward pass, in contrast to pair-wise re-ranking, which requires $k$ passes. 
To quantitatively assess the efficacy of our method, we present runtime performance analysis in Table~\ref{tab:runtime}. 
For models based on the Longformer architecture, even the largest \MODEL-base with 111M parameters exhibits latency close to that of RRT~\cite{tan2021instancelevel}. We can trade off between latency and accuracy with our three model variants. For instance, \MODEL-small offers 3$\times$ lower latency while only marginally sacrificing performance metrics on the $\mathcal{R}$Oxf and $\mathcal{R}$Par datasets on DELG descriptors.
It still offers 2$\times$ lower latency when competing with the efficiency-centric AMES~\cite{suma2024ames} method.
All variants of \MODEL exhibit significantly reduced peak memory compared to RRT, with \MODEL-tiny leading at nearly 10$\times$ lower peak memory usage.
Long-context models benefit from efficient utilization of the memory hierarchy in modern hardware accelerators, and as a result, all variants of \MODEL achieve higher throughput and increased parallelism as measured in FLOPs per second. 
\section{Conclusion}
We present \MODEL, the first image re-ranking framework that leverages list-wise re-ranking supervision at the local descriptor level. With a long-context sequence model, this approach effectively captures dependencies between the query image and each gallery image in addition to the dependencies among the gallery images themselves and implicitly learns to calibrate predictions for a more precise ranking list. 
As demonstrated through extensive ablation studies, learning from list-wise re-ranking signals enables the model to better leverage the initial rankings obtained from global retrieval.  
We achieve leading re-ranking results across established image retrieval datasets and obtain state-of-the-art metrics on the $\mathcal{R}$Oxf and $\mathcal{R}$Par datasets under the same settings.

\vspace{0.05in}
\noindent{\bf Acknowledgements.} This work was partially supported by NSF Award \#2201710 and funding from the Ken Kennedy Institute at Rice University. 
This work was also supported by the Junior Star GACR GM no. 21-28830M, Horizon MSCA-PF no. 101154126, and the Czech Technical University in Prague no. SGS23/173/OHK3/3T/13.
This work was supported in part by the NSF Campus Cyberinfrastructure grant “CC* Compute: Interactive Data Analysis Platform” NSF OAC-2019007, and by Rice University’s Center for Research Computing (CRC).

\clearpage
\setcounter{page}{1}

\maketitlesupplementary

\renewcommand\thesection{\Alph{section}}
\renewcommand{\thefigure}{\Alph{figure}}
\renewcommand{\thetable}{\Alph{table}}

\setcounter{section}{0}
\setcounter{table}{0}
\setcounter{figure}{0}

\section{Implementation details}
\label{app:model_card}

\renewcommand{\thefootnote}{\arabic{footnote}}

All training is conducted on 8 NVIDIA A100 PCI-E 40GB GPUs. 
Training on Google Landmark v2 clean set~\cite{Weyand_2020_CVPR} takes 106 hours on \MODEL-base for 5 epochs.
Models are trained with AdamW optimizer~\cite{DBLP:conf/iclr/LoshchilovH19}, $5$e$-5$ learning rate and weight decay disabled. 
Global batch size is set to 128 with 4 gradient accumulation steps. 
We present the configurations of the different \MODEL variants in Table~\ref{tab:model_card}. 
\MODEL-tiny is initialized from \texttt{roberta-tiny-cased}\footnote{\url{https://huggingface.co/haisongzhang/roberta-tiny-cased}} by migrating weights and repeatedly copying absolute position embedding along the sequence dimension\footnote{\url{https://github.com/allenai/longformer/blob/master/scripts/convert_model_to_long.ipynb}}. 
\MODEL-small is initialized from the first 6 layers of \texttt{longformer-base-4096}\footnote{\url{https://huggingface.co/allenai/longformer-base-4096}}, while \MODEL-base is initialized from the full \texttt{longformer-base-4096}.
To accommodate 50 descriptors $\times$ (1 query image + 100 re-ranking candidates) = 5,050 tokens, the position embeddings in the original models are linearly interpolated to extend their length from 4,096 to 5,120. 

When experimenting with local descriptors from DINOv2~\cite{oquab2023dinov2}, we use the same training set as AMES~\cite{suma2024ames}, which is approximately half the size of the full GLDv2 clean set, \ie 750k images.
We adopt the same global-local ensemble scheme as AMES. The ensemble hyper-parameters 
are selected based on the best-performing configuration on GLDv2 public validation split and applied to $\mathcal{R}$Oxf and $\mathcal{R}$Par evaluations. For the training of AMES$^\ast$, we follow the original training process from AMES, changing only the batch size and learning late to 150 and $1$e$-5$, respectively.

For baseline results on CUB-200~\cite{wah_branson_welinder_perona_belongie_2011}, Stanford Online Products (SOP)~\cite{song2016deep} and In-shop~\cite{liuLQWTcvpr16DeepFashion}, we reproduce them using their official code releases and identical training configuration, except for ProxyNCA++~\cite{teh2020proxynca}, we change the training image size from 256 $\times$ 256 to 224 $\times$ 224 to use the training image size same as the other baselines. 

For the performance benchmark in Section 4.3, we use the Deepspeed~\cite{deepspeed} profiler on a single NVIDIA A100 GPU to measure key performance metrics of the model per 100 re-ranked gallery images as follows: the number of parameters (\# of Params), floating-point operations (FLOPs), throughput in FLOPs per second, latency, and peak memory consumption. All dynamic metrics are reported with 10 warmup steps followed by 10 measurements for reporting the mean and standard deviation. Parameters of visual backbones are excluded from \# of Params.

\begin{table}[t]
\centering

\scalebox{0.9}{
\begin{tabular}{lccccc}
    \toprule
   \textbf{Model Variants} & \textbf{tiny} & \textbf{small} & \textbf{base} \\ 
   \midrule
   \textbf{Number of Parameters} & 19.4M & 58.7M & 111.8M \\ 
   \textbf{Number of Layers} & 4 & 6 & 12 \\ 
   \textbf{Local Attention Window} & 1024 & 512 & 512 \\ 
   \textbf{Hidden Size} & 512 & 768 & 768 \\ 
   \textbf{Intermediate Size} & 2048 & 3072 & 3072 \\ 
   \textbf{Number of Attention Heads} & 8 & 12 & 12 \\ 
   \textbf{Max Context Length} & 5120 & 5120 & 5120 \\ 
   \bottomrule
\end{tabular}
}
\caption{\textbf{Architectural parameters} of \MODEL variants.}
\label{tab:model_card}
\end{table}

We consider the descriptors to be already extracted and exclude I/O from measuring memory, latency, \etc. 
For the geometric verification (GV) method, we run RANSAC in OpenCV~\cite{itseez2015opencv} with 1,000 iterations on AMD EPYC 9354 CPU and measure the wall-clock time as the latency and the maximum resident set size (Max RSS) as the peak memory consumption.
All models are benchmarked with batched input except CVNet Reranker~\cite{DBLP:conf/cvpr/LeeS0K22}. It is worth noting that CVNet Reranker does not support batched inference since it computes pair-wise multi-scale correlation on raw feature maps (without resizing) from query and gallery images of different sizes.
Thus, CVNet Reranker heavily underutilizes the GPU and achieves extremely low throughput and high latency. The FLOP, latency, and peak memory are measured assuming query and gallery images of 512 $\times$ 512 size in CVNet Reranker. 

\begin{table*}[h]
\centering
\scalebox{1.}{
    \begin{tabular}{c @{\hspace{10pt}} c c @{\hspace{10pt}} c c c c c c}
    \toprule
    \multirow{2}{*}{\bf \shortstack{Global}} & \multirow{2}{*}{\bf \shortstack{Local}} & \multirow{2}{*}{\bf Re-rank} & \multicolumn{3}{c}{\bf $\mathcal{R}$Oxf+1M} & \multicolumn{3}{c}{\bf $\mathcal{R}$Par+1M} \\ 
       & & & \small Medium & \small Hard & \small Hard$^\star$ & \small Medium & \small Hard & \small Hard$^\star$ \\
    \midrule
       \multirow{14}{*}{SG} & \multirow{4}{*}{N/A}  &  N/A\rlap{$^\dagger$}  & \num{78.8} & \multicolumn{2}{c}{\num{61.9}} & \num{83.9} & \multicolumn{2}{c}{\num{69.1}} \\
       & &  N/A  & \num{78.5} & \multicolumn{2}{c}{\num{61.4}} & \num{83.6} & \multicolumn{2}{c}{\num{68.4}} \\
       & & SG-Rerank\rlap{$^\dagger$} & \num{84.4} & \num{71.1} & N/A & \num{84.9} & \num{71.4} & N/A \\
       & & SG-Rerank  & \num{84.0} & \num{69.4} & \num{63.9} & \num{85.2} & \num{72.3} & \num{75.7} \\  
       
       \cmidrule[0.1pt]{2-9}
       & \multirow{5}{*}{CVNet} &  R2Former  & \num{79.9} & \multicolumn{2}{c}{\num{63.7}} & \num{83.8} & \multicolumn{2}{c}{\num{69.7}} \\
       & &  RRT  & \num{79.3} & \multicolumn{2}{c}{\num{62.7}} & 83.6 & \multicolumn{2}{c}{\num{69.1}} \\
       & &  AMES  & \num{80.7} & \multicolumn{2}{c}{\num{65.7}} & \num{84.6} & \multicolumn{2}{c}{\num{71.8}} \\
       & &  \MODEL  & \num{81.9} & \num{68.6} & \num{64.9} & \num{84.6} & \num{71.4} & \num{70.7} \\
      & &   SG + \MODEL  &  \num{84.7} &  \num{71.5} &  \num{65.6} &  \num{86.2} &  \num{74.8} &  \num{76.1} \\
       \cmidrule[0.1pt]{2-9}
       & \multirow{5}{*}{DINOv2} &  R2Former$^\ast$  & \num{81.0} & \multicolumn{2}{c}{\num{66.2}} & \num{84.9} & \multicolumn{2}{c}{\num{72.1}} \\
       & &  RRT$^\ast$  & \num{81.0} & \multicolumn{2}{c}{\num{66.1}} & \num{85.5} & \multicolumn{2}{c}{\num{73.3}} \\
       & &  AMES$^\ast$  & \num{81.3} & \multicolumn{2}{c}{\num{67.3}} & \num{85.8} & \multicolumn{2}{c}{\num{74.3}} \\
       & &  \MODEL  & \num{85.8} & \num{75.8} & \num{73.2} & \num{86.8} & \num{75.9} & \num{76.5} \\  
       & &   SG + \MODEL  &  \num{86.5} &   \num{76.3} &  \num{73.7} &  \num{87.2} &   \num{76.9} &  \num{78.2} \\
    
    \midrule
    
    \multirow{7}{*}{DINOv2} & \multirow{2}{*}{N/A} & N/A & \num{59.6} & \multicolumn{2}{c}{\num{35.2}} & \num{77.0} & \multicolumn{2}{c}{\num{58.9}} \\ 
     & & SG-Rerank  & \num{62.2} & \num{40.5} & \num{31.2} & \num{79.8} & \num{60.5} & \num{65.8} \\ \cmidrule[0.1pt]{2-9}
     & \multirow{5}{*}{DINOv2} &  R2Former$^\ast$  & \num{67.8} & \multicolumn{2}{c}{\num{44.6}} & \num{78.6} & \multicolumn{2}{c}{\num{61.3}} \\
     & & RRT$^\ast$  & \num{68.8} & \multicolumn{2}{c}{\num{46.0}} & \num{79.6} & \multicolumn{2}{c}{\num{64.0}} \\
     & & AMES$^\ast$  & \num{68.9} & \multicolumn{2}{c}{\num{46.8}} & \num{79.9} & \multicolumn{2}{c}{\num{64.7}} \\
     & & \MODEL &  \num{73.4} & \num{54.9} & \num{52.5} & \num{80.9} & \num{66.4} & \num{66.7} \\
     & &   SG + \MODEL  &  \num{71.2} & \num{54.4} &  \num{48.7} & \num{81.9} &  \num{68.7} &  \num{69.5} \\
    \bottomrule
    \end{tabular}
}
    \caption{\textbf{Additional results} with re-ranking top-400 candidates. 
    Hard$^\star$:  {\em easy} images are completely removed from the database.
    Hard:  {\em easy} images are used to re-rank and removed before the evaluation.
    $\dagger$: results in the SuperGlobal paper~\cite{Shao_2023_ICCV}. \MODEL is reported with the base variant. SG + \MODEL: re-ranking with SG first and then with \MODEL. $^\ast$ indicates models trained with 768 hidden size, serving as a fair comparison with LOCORE. N/A: not available.}
    \label{tab:table1_ext}
\end{table*}

\section{Additional Experimental Results}
\label{app:additional_results}
\subsection{Additional comparisons}

We present additional experiments with different combinations of global and local features in Table~\ref{tab:table1_ext}. We compare with more baseline re-ranking methods, including methods with global, \ie SuperGlobal (SG) Rerank~\cite{Shao_2023_ICCV}, and local, \ie AMES~\cite{suma2024ames}, RRT~\cite{tan2021instancelevel}, R2Former~\cite{r2former}, descriptors. We evaluate the models under different Hard settings, using different global descriptors to generate the shortlist and different backbones for feature extraction. We also test the combination of \MODEL with other re-ranking schemes.

\textbf{Variations for Hard setup.} 
As mentioned in the main paper, there can be two approaches regarding how to handle {\em easy} images in the hard setup: (i) \textbf{Hard}: {\em easy} images are used to re-rank and removed before the evaluation (typically used in the literature~\cite{Shao_2023_ICCV}), and (ii) \textbf{Hard$^\star$}: {\em easy} images are completely removed from the database.
While the two choices (Hard and Hard$^\star$) are equivalent for pair-wise re-ranking methods, this is not the case when interactions between database images are considered (\ie \MODEL, SG-rerank). In Table~\ref{tab:table1_ext}, it is evident that the two setup lead to significantly different results. In most cases, mAP considerably drops in $\mathcal{R}$Oxf, comparing results in Hard and Hard$^\star$; whereas, mAP increases in $\mathcal{R}$Par.

\textbf{Performance with other backbones.} 
First, we benchmark all models when the shortlist is generated based on DINOv2 global descriptors. It is noteworthy that DINOv2 global descriptors are significantly worse than SG ones. In this setup, \MODEL outperforms all other re-ranking schemes by a vast margin. 

Second, we evaluate \MODEL using local descriptors extracted from CVNet backbones. CVNet local descriptors have a higher dimension than that of DINOv2, \ie 1024 vs 768; hence, we used a learnable linear projector to match the embedding dimensionality of the transformer. \MODEL achieves competitive performances compared with the pair-wise re-rankers, with only AMES outperforming it in a few cases. Yet, all local-based re-rankers are outperformed by SG-Rerank. Nevertheless, \MODEL with DINOv2 outperforms SG-Rerank.

\textbf{Combination with SG-Rerank.} 
It is straightforward to combine local and global-based re-ranking. To this end, we combine \MODEL with SG-Rerank by applying global re-ranking first, followed by local re-ranking. This combination achieves the best performance when SG is used as global descriptor. However, this combination hurts \MODEL performance on $\mathcal{R}$Oxf when DINOv2 is used as global.

\begin{figure*}[t]
	\begin{center}
    \includegraphics[width=0.96\linewidth]{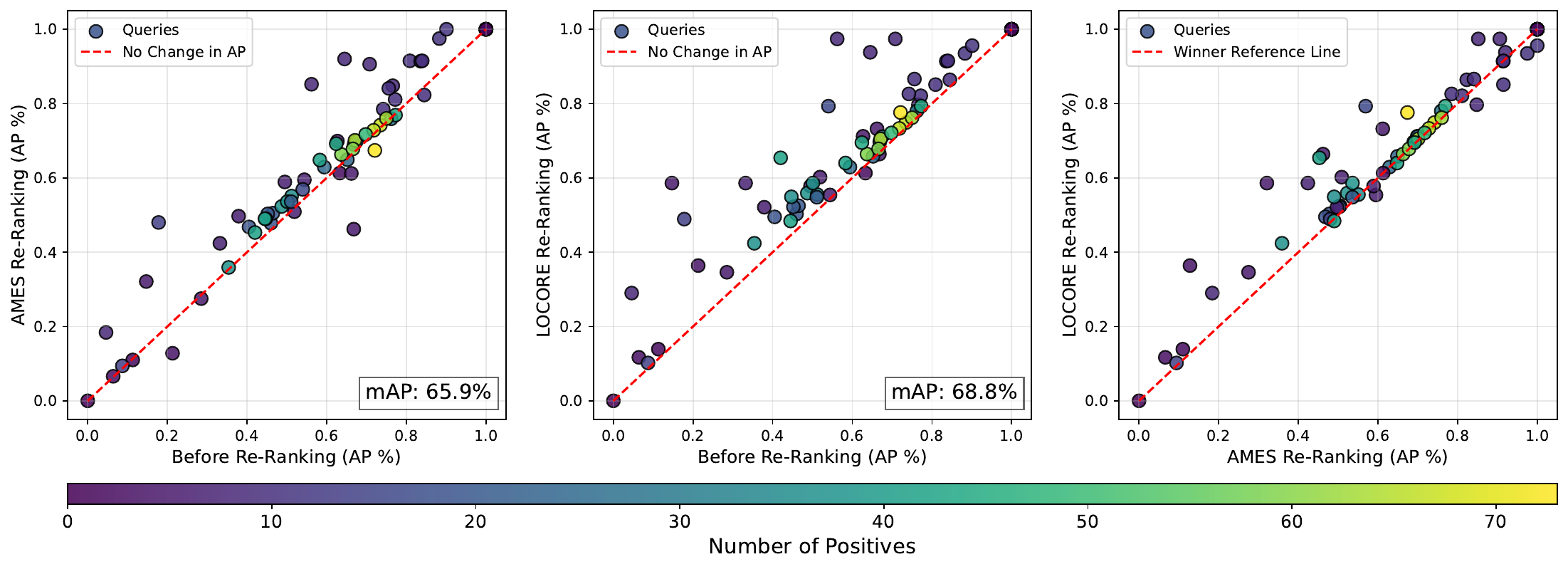}
	\end{center}
    \vspace{-15pt}
	\caption{\textbf{Average precision per query scatter plot} on $\mathcal{R}$Oxf+1M Hard for global-only vs. AMES (\textit{Left}), global-only vs. \MODEL-small (\textit{Middle}) and AMES vs. \MODEL-small (\textit{Right}). Global descriptors are from RN101-Superglobal, which by itself achieves mAP=61.4\%. Re-ranking is performed for top-100 candidates, and the color bar indicates the number of positive images in the shortlist for each query.}
	\label{fig:analysis}
    \vspace{-10pt}
\end{figure*}

\textbf{Performance per query.} 
To highlight the advantages of our proposed list-wise re-ranking over pair-wise re-ranking, we present several scatter plots in Figure~\ref{fig:analysis}, showing the average precision of each sample in $\mathcal{R}$Oxf+1M Hard before and after re-ranking with different re-ranking paradigms. We compare our model with AMES~\cite{suma2024ames}, which is considered the state-of-the-art solution for pair-wise re-ranking.
In the first two plots, we observe that most data points are concentrated in the upper-left half and above the red reference line, indicating that both re-ranking paradigms improve the ranking accuracy for the majority of query images.
However, the list-wise re-ranking method driven by \MODEL has barely any sample points below the red reference line, meaning the re-ranking only improves the retrieval on the individual query level. 
The distinction between the two models is most prominent in the final plot, where the number of sample points above the winner reference line far exceeds those below, demonstrating that \MODEL outperforms AMES on more query samples.
We also observed that the list-wise re-ranking method is relatively robust in terms of the number of positive samples included in the shortlist, as the color distribution of the sample points does not exhibit any discernible pattern. This indicates the general versatility of \MODEL.

\begin{table}[t]
\small
\setlength{\tabcolsep}{7pt}
\centering
\scalebox{0.8}{
    \begin{tabular}{l @{\hspace{10pt}} l r r @{\hspace{10pt}} c c c c c c}
        \toprule
        \multirow{2}{*}{\bf \shortstack{Global}} & \multirow{2}{*}{\bf \shortstack{Local}} & \multirow{2}{*}{$L$} & \multirow{2}{*}{$K$} & \multicolumn{2}{c}{\bf $\mathcal{R}$Oxf+1M} & \multicolumn{2}{c}{\bf $\mathcal{R}$Par+1M} \\ 
           & & & & \small Medium & \small Hard$^\star$ & \small Medium & \small Hard$^\star$ \\
        \midrule
           \multirow{3}{*}{SG} & \multirow{3}{*}{DINOv2}  & \num{50} & {100}  & \num{85.8} & \num{73.2} & \num{86.8} & \num{76.5} \\ 
           & & 100 & 50 & \num{83.9} & \num{68.6} & \num{85.2} & \num{72.5} \\ 
           & & 25 & 200 & \num{84.5} & \num{72.1} & \num{85.6} & \num{75.1} \\ 
        \bottomrule
    \end{tabular}
}
\vspace{-5pt}
\caption{Additional results for \MODEL-base with different combinations of the number of local descriptors $L$ and the number of re-ranking candidates $K$ on $N=400$ candidates.
}
\vspace{-5pt}
\label{tab:f_and_r}
\end{table}

\subsection{Additional ablations}

\textbf{Number of images vs number of descriptors.} 
We explore the relationship between the number of local descriptors and the number of image candidates within a given context window in Table~\ref{tab:f_and_r}. 
Specifically, we set the context window to 5,120 and examine three configurations of \MODEL: (i) using 100 gallery images with 50 local descriptors per image, \ie the default setup, (ii) using 200 gallery images with 25 local descriptors per image, \ie more candidate images but fewer descriptors per image, and (iii) using 50 database images with 100 local descriptors per image, \ie more descriptors per image but fewer candidate images. The \MODEL in the default settings reports the best results.

\begin{table}[t]
\centering
\small
\setlength{\tabcolsep}{8pt}
\renewcommand{\arraystretch}{1.1}
\begin{tabular}{l @{\hspace{10pt}} lll}
\toprule
\textbf{Ablation Module} & \hspace{-4pt} $\mathcal{R}@1$ & \hspace{-7pt} $\mathcal{R}@10$ & \hspace{-11pt} mAP@$R$ \\
\midrule
Global descriptors & 80.8 & 92.1 & 65.1 \\
\midrule
\MODEL-tiny & 82.4 & 93.1 & 68.0 \\
\MODEL-small & 83.3 & 92.7 & 69.4 \\
\MODEL-base & 83.8 & 92.9 & 71.0 \\
\midrule
\MODEL-RWKV & 81.4 & 92.3 & 66.7 \\
\MODEL-Mamba & 80.6 & 92.1 & 66.4 \\
\bottomrule
\end{tabular}
\vspace{-5pt}
\caption{Ablation studies for \MODEL recurrent models on the SOP dataset. Re-ranking is performed with the top 100 candidates.
}
\label{tab:ablation_sop}
\vspace{-10pt}
\end{table}

\textbf{Comparison with other recurrent models.}
Other model architectures with no restrictions on context length that could be employed instead of LongFormer are the recently proposed recurrent models Mamba~\cite{gu2023mamba} and RWKV~\cite{peng2023rwkv}. As the causal nature of the recurrence-based model does not align well with our re-ranking motivation and is strictly less expressive than bi-directional encoders~\cite{DBLP:conf/icml/JiaYXCPPLSLD21, DBLP:conf/icml/RadfordKHRGASAM21}, we follow the common practice in recurrent visual encoder community~\cite{duan2024visionrwkv, li2024videomamba, liu2024vmamba} to build a bi-directional variant that serves as an efficient sequence encoder. 
To ensure recurrent models can still handle long-range interactions and alleviate the inherent information bottleneck in the design of recurrent models, we devise a mechanism that resembles the query global attention in Section 3.2 by interleaving recurrent blocks with uni-directional transformer blocks~\cite{vaswani2017attention}. 
These transformer blocks compute attention scores between intermediate hidden states of query image tokens and intermediate hidden states of gallery image tokens and produce fused intermediate representations for the following layers to process. The uni-directional attention guarantees that every gallery image has similar difficulty accessing the query image, irrespective of its position in the sequence relative to the query. 
Although we find that these recurrence-based models could slightly outperform the base global retrieval model, they do not surpass our transformer-based results, as shown in Table~\ref{tab:ablation_sop}.

\begin{figure*}[ht]
	\begin{center}
		\includegraphics[width=\linewidth]{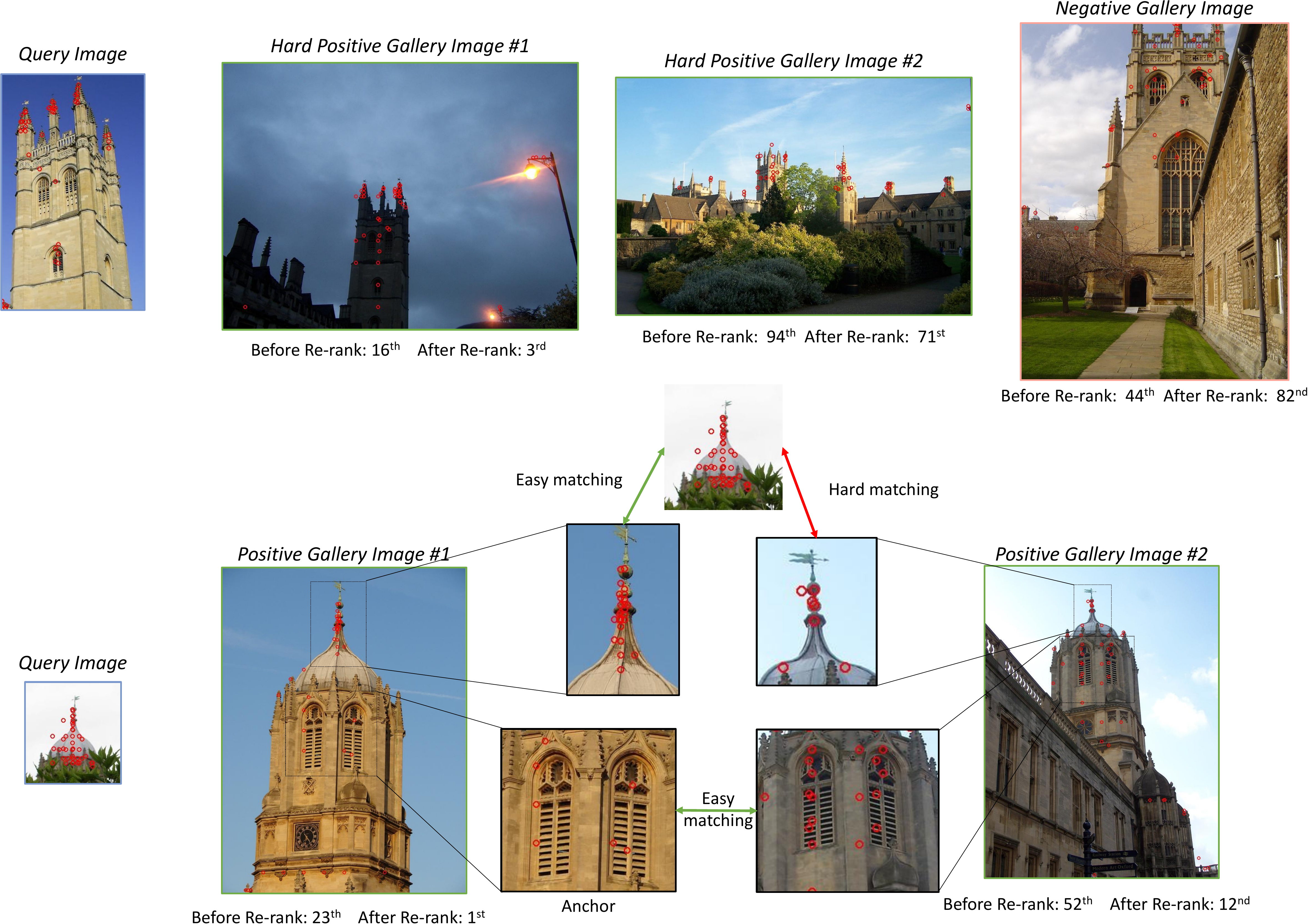}
	\end{center}
	\caption{\textbf{Qualitative analysis} on $\mathcal{R}$Oxford dataset of \MODEL-base on RN50-DELG descriptors. \textbf{Upper:} two hard positive gallery images get assigned with higher ranks while a negative gallery image is put in lower ranks after re-ranking. \textbf{Lower:} the first gallery image can be easily identified as positive due to its dense matching with the query image; it can also serve as a perfect anchor image for refining the ranking of the second gallery image due to their transitive relationship. }
	\label{fig:qualitative}
\end{figure*}

\begin{figure}[t]
  \centering
  \includegraphics[width=\linewidth]{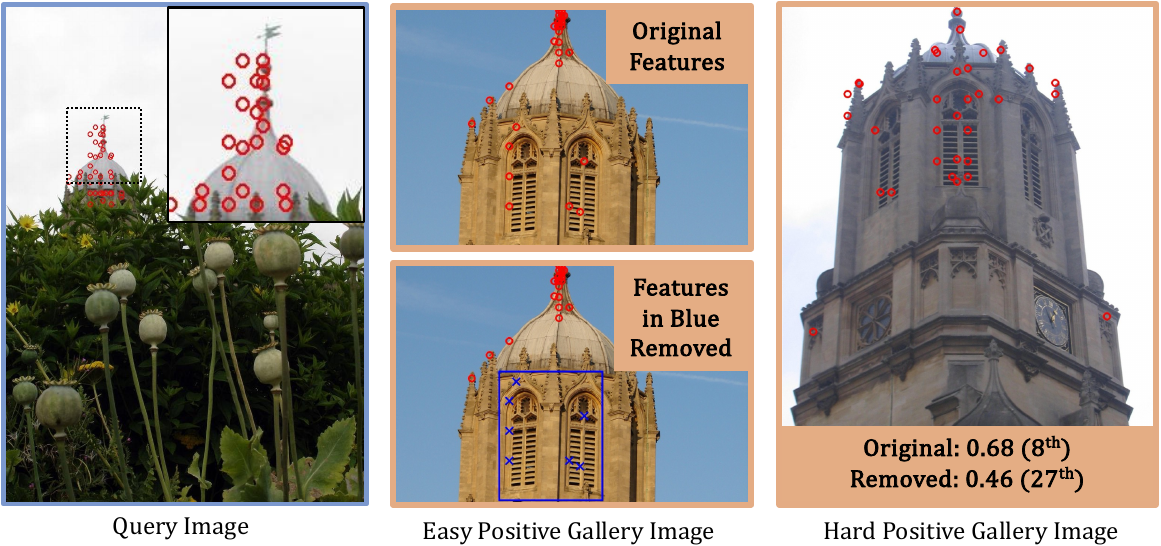}
  \caption{\textbf{Visualization of \MODEL capturing transitive relationships in gallery images.}
  We prevent \MODEL from accessing local features of the easy positive corresponding to the windows (blue crosses) and instead randomly sample local features from other negative images. 
  The dropped similarity score indicates \MODEL relies on the transitivity of local features to calibrate predictions for hard positive gallery images.
  }
  \label{fig:supp_example}
\end{figure}


\subsection{Qualitative Results}
\label{app:more_qualitative}

We illustrate the re-ranking performance of \MODEL in Figure~\ref{fig:qualitative} as qualitative results. The upper example underscores the superior performance of our method, demonstrated by its success in elevating the ranking of two hard positive images and lowering that of the negative gallery image. 
We also show in the lower example that our model is able to capture the transitive relationship between query and gallery images. 
The transitive relationship is based on the assumption that generally, if two gallery images are similar and one of them is predicted as positive, then the other should be calibrated with higher confidence. 
In the lower example, the correspondence between the query image and the first gallery image is easy to catch as the common geometric features are evident, resulting in Easy matching in the figure. 
However, although the global retriever returns the second gallery image as re-ranking candidates, the sparse local features focused on the top of the tower make it hard for pair-wise re-ranker to assign this gallery image a high confidence score. This misalignment is calibrated by our list-wise re-ranking paradigm since the windows in both gallery images can serve as the anchor to propagate the positive prediction from the easy candidate to the hard one.

Additionally, in Figure~\ref{fig:supp_example} the easy positive gallery has visual overlap with the query (rooftop). The hard positive gallery has little visual overlap with the query, but larger overlap with the first positive (\eg windows). We wish to answer this question: \textit{Are the local features of the window improving the rank of the hard positive due to a transitive relationship? }
We remove local features of the windows (blue crosses), repeat the similarity estimation, and compare the ranks. The decreased similarity score is a sign of \MODEL capturing transitive relationships.

\section{Limitations and Future Work}
\label{sec:limitations}
Despite the merits in efficiency and re-ranking performance, our model is inherently restricted by the context window of existing encoder-only sequence models. 
A limited context window limits the number of re-ranking candidates in the gallery and the number of local descriptors that \MODEL can use. 
While recurrent models offer more flexibility with the context window size, we find that they could not capture list-wise re-ranking dependencies as well as transformer-based models, resulting in sub-optimal performance. 
Future work could adopt large-scale decoder-only sequence models which typically have longer context windows and greater capacity for list-wise re-ranking.
Additionally, context parallelization techniques (\eg, RingAttention~\cite{liu2023ring}, Infini-attention~\cite{munkhdalai2024leave}) could help expand the context window of current Transformer encoder models. 
Lastly, extractive re-ranking as proposed in our work could also be seamlessly adopted for other modalities, \eg document or video re-ranking. 

{
    \small
    \bibliographystyle{ieeenat_fullname}
    \bibliography{main}
}

\end{document}